\documentclass{article}
\usepackage{spconf,amsmath,graphicx}
\usepackage{booktabs}
\usepackage[table,xcdraw]{xcolor}

\usepackage{enumitem}
\setlist{nosep, leftmargin=14pt}

\usepackage{mwe} 
\usepackage{color}
\usepackage{url}
\usepackage{comment}

\newcommand{\wqz}[1]{\textcolor{black}{#1}}

\title{Video4MRI: An Empirical Study on Brain Magnetic Resonance Image Analytics with CNN-based Video Classification Frameworks}
\name{Yuxuan Zhang$^{\dagger1,2}$, Qingzhong Wang$^{\dagger1}$, Jiang Bian$^{\dagger1}$, Yi Liu$^1$, Yanwu Xu$^1$, Dejing Dou$^1$, Haoyi Xiong$^{*1}$\thanks{$\dagger$ Equal contribution. The work was done when Yuxuan Zhang was an intern at Baidu. $^*$Correspondence to  Haoyi Xiong via haoyi.xiong.fr@ieee.org.}}
\address{$^1$~Baidu Inc., Beijing, China. \\
$^2$~School of Data Science, The Chinese University of Hong Kong, Shenzhen, China.
}
\begin{document}
%
\maketitle
\begin{abstract}
\wqz{To address the problem of medical image recognition, computer vision techniques like convolutional neural networks (CNN) are frequently used. Recently, 3D CNN-based models dominate the field of magnetic resonance image (MRI) analytics. Due to the high similarity between MRI data and videos, we conduct extensive empirical studies on video recognition techniques for MRI classification to answer the questions: (1) can we directly use video recognition models for MRI classification, (2) which model is more appropriate for MRI, (3) are the common tricks like data augmentation in video recognition still useful for MRI classification? Our work suggests that advanced video techniques benefit MRI classification.} 
In this paper, four datasets of Alzheimer's and Parkinson's disease recognition are utilized in experiments, together with three alternative video recognition models and data augmentation techniques that are frequently applied to video tasks. In terms of efficiency, the results reveal that the video framework performs better than 3D-CNN models by $5\%\sim11\%$ with $50\%\sim66\%$ less trainable parameters. This report pushes forward the potential fusion of 3D medical imaging and video understanding research.

\end{abstract}

\vspace{-2mm}
\section{Introduction}\vspace{-2mm}
\label{sec:intro}
Magnetic Resonance Imaging (MRI) is a common diagnostic tool established for the identification of brain lesions caused by neurological diseases, such as Alzheimer’s Disease (AD), Parkinson’s Disease (PD), etc. Early identification is crucial for these diseases since treatment can halt the progression and lessen symptoms. Degenerative illnesses like AD and PD, in contrast to brain tumors, cause age-related brain deterioration throughout the entire brain rather than just one specific area. For instance, the gyri may shrink, the sulci may deepen, and the cerebral cortex may grow thinner~\cite{8211486}. In clinical practice, the indicators of brain degeneration are manually identified by physicians and doctors. However, such identifications would be challenging when clinical staffs are not well-experienced or the patient’s case is not typical. 

With the development of deep neural networks~\cite{lecun2015deep}, 
data-driven medical image analysis experiences rapid growth, 
serving as auxiliary tools for physicians to 
reduce the probability of misdiagnosis and \wqz{speed up diagnosis processes}. Previous studies have demonstrated that recognition pipelines developed for natural 
images 
can be transferred~\cite{li2020rifle} to medical image classification 
with good performance. 
For 3D MRI classification 
most well-implemented frameworks are based on 3D CNNs, which extracts feature in 3 axes. However, compared with 2D networks, 3D networks have more parameters and higher computational complexity, 
slow down the training process. \wqz{Researchers in the field of video recognition have proposed many solutions to effectively extract both space and time features, speeding up training and inference processes, and boosting the performance.}
The goal of this work is to determine whether architectures designed for video applications can efficiently and effectively extract both plane and axial spatial features from 3D MRI for the purpose of identifying diseases or diagnosing.

\begin{figure}[t]

  \centering
  \centerline{\includegraphics[width=0.8\linewidth]{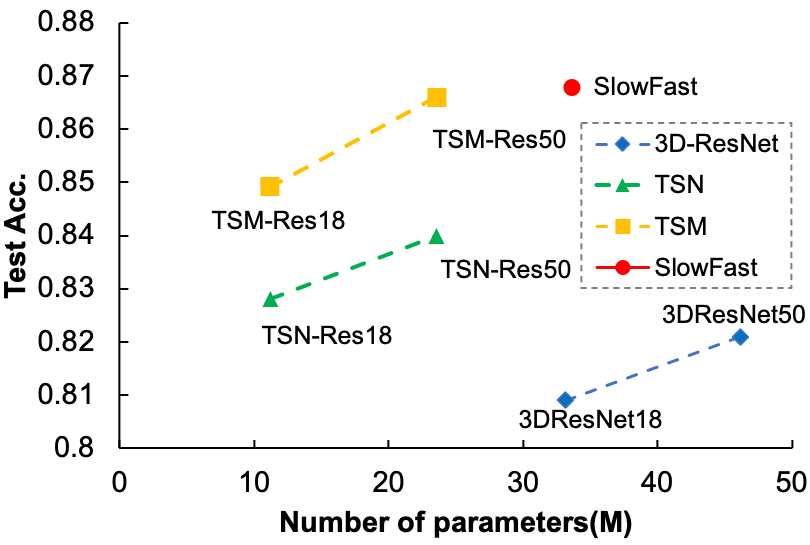}}
%
\vspace{-3mm}
\caption{Test accuracy vs. the number of parameters of different models. We test all models on the PPMI dataset.
} \label{fig1}
\vspace{-3mm}
\end{figure}

In this work, empirical studies of video recognition frameworks for MRI classification are conducted. 
\wqz{Specifically, we compare 3D and advanced 2D models on} 
4 MRI datasets with different diseases (AD and PD) and 
views (sagittal and axial). \wqz{To test the ability of different models and for fair comparisons, we train all models from scratch using raw magnetic resonance images, i.e., model parameters are randomly initialized and input images are not pre-processed, keeping the same as video recognition.}
Our experimental results show that \wqz{(1) directly applying video recognition frameworks to MRI classification is able to achieve satisfying performance, i.e., Temporal Shift Module (TSM) \cite{lin2019tsm} with ResNet-50 \cite{he2016deep} as the backbone achieves 86.6\% accuracy on PPMI dataset \cite{marek2011parkinson}, (2) advanced 2D models are more efficient and effective than the widely applied 3D models, e.g., TSM with less training time and fewer parameters obtains better performance on 4 datasets than 3D models. Fig.~\ref{fig1} presents a comparison among different models in the perspective of performance and the number of model parameters, (3) data augmentations which are widely used in video recognition benefit MRI classification, e.g., using data augmentation remarkably improve the performance by around 10\%.}

\vspace{-2mm}
\section{Preliminary 
}\vspace{-2mm}
\label{sec:format}

\wqz{In this section, we present a brief overview of Brain MRI tasks and video analytical models.}

\vspace{-3mm}
\subsection{Brain MRI classification}\vspace{-2mm}
\wqz{The task of brain MRI classification requires a model to provide a label that indicates whether the patient is healthy given a set of magnetic resonance images.}
Over the past decade, CNN-based 
models are widely used in dozens of studies on brain disorders like Parkinson's disease\cite{9288068,zhang2022mining}, Alzheimer's disease\cite{kushol2022addformer,li2022trans,dhinagar20213d,liang2021alzheimer}, Schezo-phrenia\cite{oh2020identifying}, etc. Those proposed models can be divided into two main categories. The first trend is 3D CNN-based models that use 3D convolutional kernels to sufficiently use voxel information in the whole MRI \cite{dhinagar20213d}. 3D methods normally require high computational costs and have high computational complexity. The second trend is 2D CNN models that process slices of MRI independently and fuse the decision of the model in late stage \cite{liang2021alzheimer}. These approaches reduce computational costs and complexity. \wqz{Also, one can easily plug pre-trained 2D backbones into classifiers, so as to accelerate their convergence during fine-tuning~\cite{xiong2022grod}.} 
However, 2D approaches neglect spatial correlation among slices, \wqz{which is crucial for classification.}

\vspace{-3mm}
\subsection{Video recognition frameworks}\vspace{-2mm}
Video 
recognition is a representative task in computer vision \cite{wu2022survey}. Compared with image recognition, the main challenge in video recognition is to efficiently extract temporal information from videos. 
A straightforward idea is to extend a 2D model to its 3D version using 3D convolutional kernels to replace 2D kernels. Representative works include I3D \cite{carreira2017quo}, R3D \cite{Hara_2017_ICCV}, etc. These architectures achieve satisfying performance but have much higher computational costs, \wqz{requiring more training time and GPU memory}. \wqz{Alternatively, some 2D models extract features from video frames or other auxiliary features like optical flow using 2D convolutional kernels. The representative works include Two-stream network\cite{simonyan2014two}, Temporal Segment Network (TSN) \cite{wang2016temporal}, etc.}
These methods reduce computational costs and complexity but they \wqz{just aggregate the frame-level labels to obtain a video-level label, which cannot capture much temporal information.}
\wqz{In contrast, TSM \cite{lin2019tsm} employs 2D convolutional kernels and designs a shift module to imitate 3D convolution, reducing computational costs and achieving better performance than 3D models. SlowFast \cite{feichtenhofer2019slowfast} applies two branches to model slow and fast motions in videos, where the slow branch uses a low sampling rate and the fast branch uses a high sampling rate. In this paper, we focus on 3 typical video recognition frameworks: \textbf{TSN}, \textbf{TSM} and \textbf{SlowFast}.} 

\textbf{TSN} 
is a 2D video recognition network. In the training process, the input video is divided into segments and \wqz{it randomly selects one frame from each segment. A 2D CNN is applied to each selected frame to extract features and finally TSN uses the mean feature of each frame for classification. In addition to frames, the optical flows of each selected frame are used to provide motion information.}

\textbf{TSM} 
\wqz{also employ 2D CNNs to extract frame-level features, but it} models temporal motion 
by shifting the feature maps on the temporal axis. Such operation does not increase computational cost but well captures temporal information in videos. \wqz{In TSM, residual networks (ResNet) are used and temporal shift performs on each residual block. Thus, temporal information can be captured in multiple stages with different scales. Finally, TSM outputs results with a linear model.}

\textbf{SlowFast} leverages a multi-path architecture~\cite{zhao2021efficient} and 
is composed of two independent paths which sample the input video at different rates.  The slow path has a lower sample rate and employs more convolutional kernels to capture semantic features, 
whereas the fast path has a higher sample rate and fewer 
convolutional kernels to capture motion features. The slow path uses both 2D and 3D convolutional kernels and the fast path is a small 3D network. Such model efficiently captures features along 3 axes simultaneously.

\vspace{-2mm}
\section{Experimental settings
}\vspace{-2mm}
\label{sec:pagestyle}

\subsection{Datasets}\vspace{-2mm}
To illustrate that video recognition frameworks are adaptive to MRI classification, the models are separately tested on 
4 datasets \wqz{for two diseases, including} 
ADNI\footnote{\url{https://adni.loni.usc.edu}}, OASIS-1, OASIS-2\footnote{\url{https://www.oasis-brains.org}} for Alzheimer's Disease (AD) and PPMI~\cite{marek2011parkinson}\footnote{\url{https://www.ppmi-info.org}} 
for Parkinson's Disease (PD). Each dataset contains brain MRIs of demented patients (with AD/PD) and Non-demented control (NC) patients. More information is shown in Table \ref{tab1}.

\begin{table}[t]
\caption{Detailed information of used datasets.}\label{tab1}
\centering
\scalebox{0.8}{
\begin{tabular}{ccccc}

\cline{1-4}
Dataset & Total subjects  & Total images & View    &  \\ \cline{1-4}
PPMI    & 97(PD):81(NC)   & 222          & sagital &  \\
ADNI    & 228(AD):188(NC) & 416          & sagital &  \\
OASIS-1 & 100(AD):136(NC) & 236          & axial   &  \\
OASIS-2 & 80(AD):86(NC)   & 1368         & sagital &  \\ \cline{1-4}
\end{tabular}
}
\vspace{-1.5em}
\end{table}

\vspace{-3mm}
\subsection{Models}\vspace{-2mm}
Three series of video models are modified for MRI tasks and used in this research: TSN, TSM, and SlowFast. The models take raw magnetic resonance images as the only input modality. Given the characteristics of MRI data, we employ the TSN sampling technique, which divides the input MRI into $K$ evenly spaced segments and randomly selects 1 frame from each segment during training. In order to acquire testing results under set conditions, the sampler takes the middle frame from each segment during the testing procedure. The TSN model in this work simply feeds $K$ frames of images with shape \wqz{$W\times H$} 
into the 2D backbone network with shared parameters and we can obtain 
a $D$-dimensional feature vector for each frame, thereby acquiring a $K \times D$ feature matrix of the input MRI. The \wqz{final} 
feature vector for classification is $D$-dimensional and is obtained by simply taking the mean of $K$ frame features. Then a fully connected classification head outputs the 
probability of 2 classes. \wqz{In our experiments, we set $K=32$, $W=224$, $H=224$ and $D=2048$. We also conduct ablation studies on $K$.}

The TSM model has the same basic structure as the TSN with the addition of a temporal-shift operator on each residual block of the backbone. Two versions of the backbone, ResNet-18, and ResNet-50 are tested in TSN and TSM. 
SlowFast model with ResNet-50 architecture 
is used, 
where the frame rate in the fast path is 8 times larger than that of the slow path in this paper, \wqz{i.e., the slow path samples 4 frames and fast path samples 32 frames.}
%
Finally, 3D-ResNet-18 and 3D-ResNet-50
are compared with the above video recognition models using the same experimental settings. 

\vspace{-3mm}
\subsection{Data Augmentation}\vspace{-2mm}
\wqz{Since deep models are normally large, leading to overfitting on small-scale datasets with hundreds of data samples. One of the most widely applied techniques to mitigate this issue is using data augmentations during training.}
\wqz{In this paper, we employ video recognition frameworks for MRI classification, hence the data augmentation strategies that are commonly used in video recognition can also be employed to alleviate the problem of overfitting, improving the performance.}
In video recognition frameworks, a training sample is composed of a sequence of images and we successively apply random resize crop, brightness transform, and random rotation with a small angle to each image as a simple augmentation. \wqz{Recently, VideoMix \cite{yun2020videomix} shows a strong ability to improve the performance of deep models. In this paper, we investigate random rotation, CutMix \cite{yun2019cutmix}, and MixUp \cite{zhang2017mixup} for MRI classification. Specifically, we rotate each image with a random angle ranging from [$-15^{\circ}, 15^{\circ}$], and randomly apply spatial CutMix and MixUp with 60\% and 40\% chances.}

\vspace{-2mm}
\subsection{Training and test}\vspace{-2mm}
All models are trained from scratch using a single Tesla V100 GPU. \wqz{The batch size is 4 to fully utilize GPU memory.}
Model parameters are initialized using Kaiming initialization \cite{he2015delving} and optimized using cross-entropy loss. The learning rate is set to $5\times10^{-5}$ after 5 warm-up epochs, then arrives at zero with a single cosine cycle. Models are trained for 100 epochs for all datasets, and in the first 80 epochs, we use mix-based augmentation. 
We also use weight decay for the final FC layer in the classification head of TSM and TSN, and 
the coefficient of weight decay is set to $10^{-4}$ and the dropout \cite{srivastava2014dropout} rate is 0.4. 

\wqz{All 4 datasets are split into the training set and test set with a ratio of 7:3 on the subject level such that images obtained from the same individual will not appear in the training set and test set simultaneously. Since the test set for each dataset is small, the final accuracy is calculated by bootstrapping the test set. Specifically, the bootstrapping algorithm samples $40\%$ of the test set with replacement and calculate accuracy for 100 times, then the mean accuracy and its $95\%$ confidence interval are used as evaluation metrics of model performance. We also tested the training time for each epoch, the total number of trainable parameters, and inference time per sample to evaluate 
the efficiency of each model.}

\begin{table*}[ht]
\centering
\begin{tabular}{cc}
\begin{minipage}[b]{0.6\linewidth}
\centering
\caption{Results of test accuracy and confidence interval on 4 datasets.}\label{tab2}

\scalebox{0.65}{
\begin{tabular}{@{}cccccccc@{}}
\toprule
Acc. (95\%CI) & PPMI                  & ADNI                  & OASIS-1               & OASIS-2               & \begin{tabular}[c]{@{}c@{}}Training \\ time(s)\end{tabular} & \begin{tabular}[c]{@{}c@{}}Params \\ number(M)\end{tabular} & \begin{tabular}[c]{@{}c@{}}Inference \\ time(ms)\end{tabular} \\ \midrule
3DResNet-18   & 0.809(0.122)          & 0.622(0.101)          & 0.752(0.143)          & 0.647(0.062)          & 36               & 33.16            & 38.4               \\
3DResNet-50   & 0.821(0.118)          & 0.581(0.108)          & 0.758(0.126)          & 0.657(0.062)          & 43               & 46.16            & 61.5               \\
TSN-Res18     & 0.828(0.108)          & \textbf{0.739(0.083)} & 0.725(0.133)          & 0.716(0.056)          & 19               & 11.19            & 28.9               \\
TSN-Res50     & 0.840(0.111)          & 0.736(0.078)          & 0.777(0.109)          & 0.689(0.060)          & 24               & 23.56            & 59.6               \\
TSM-Res18     & 0.849(0.103)          & 0.738(0.094)          & 0.753(0.131)          & \textbf{0.752(0.054)} & 20               & 11.19            & 28.9               \\
TSM-Res50     & 0.866(0.102)          & 0.732(0.089)          & 0.794(0.106)          & 0.718(0.057)          & 26               & 23.56            & 59.6               \\
SlowFast      & \textbf{0.868(0.107)} & 0.737(0.095)          & \textbf{0.809(0.101)} & 0.730(0.050)          & 38               & 33.61            & 87.9               \\ \bottomrule
\end{tabular}
}
\end{minipage}
& \begin{minipage}[b]{0.38\linewidth}
\centering
\caption{Ablation study on data augmentation}\label{tab3}
\scalebox{0.73}{
\begin{tabular}{@{}ccc@{}}
\toprule
                        & PPMI                 & ADNI                  \\ \midrule
TSM                     & 0.793(0.121)         & 0.630(0.092)          \\
TSM+rotate              & 0.832(0.106)         & 0.659(0.096)          \\
TSM+MixUp/CutMix        & 0.810(0.117)         & 0.730(0.092)          \\
TSM+rotate+MixUp        & 0.813(0.110)         & 0.516(0.106)          \\
TSM+rotate+CutMix       & 0.853(0.114)         & \textbf{0.732(0.089)} \\
TSM+rotate+MixUp/CutMix & \textbf{0.866(0.102)} & 0.708(0.087)          \\ \bottomrule
\end{tabular}
}
\end{minipage}
\end{tabular}
\end{table*}

\vspace{-2mm}
\section{Results and Discussions}\vspace{-2mm}
\label{sec:typestyle}



Table \ref{tab2} shows the results of all models on 4 datasets. 
\wqz{Obviously, directly applying advanced video recognition frameworks to MRI classification can achieve satisfying performance. Compared to the most widely used 3D models in MRI classification, advanced video recognition frameworks show superiority in both effectiveness and efficiency. For example, SlowFast achieves 0.868 accuracy on PPMI, 5.72\% higher than that obtained by 3DResNet-50 (0.868 vs. 0.821), and the training time decreases by 11.6\% (33.61 vs. 46.16). TSN and TSM are much faster in both training and inference phases, e.g., the inference time of TSM-Res18 is 24.4\% less than that of 3DResNet-18 (28.9 vs. 38.4) and around 1/3 of SlowFast (28.9 vs. 87.9). In terms of the effective performance, TSM-Res50 is superior to 3D ResNet on all datasets and the improvement ranges from 4.75\% to 16.23\%. Though TSM and TSN are slightly inferior to SlowFast on PPMI and OASIS-1, they have fewer learnable parameters and run faster.} 
\wqz{One possible reason for the superiority of 2D models is that they use 2D kernels to extract plane features frame by frame, which is similar to manual diagnosis by physicians.}

\vspace{-2mm}
\subsection{Ablation study the number of extracted frames}
\begin{figure}[t]
\centering
  \includegraphics[width=0.48\linewidth]{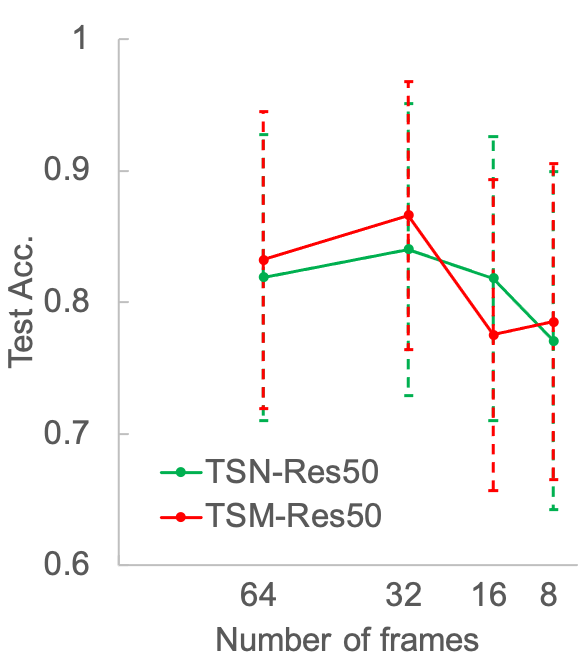}
  \includegraphics[width=0.48\linewidth]{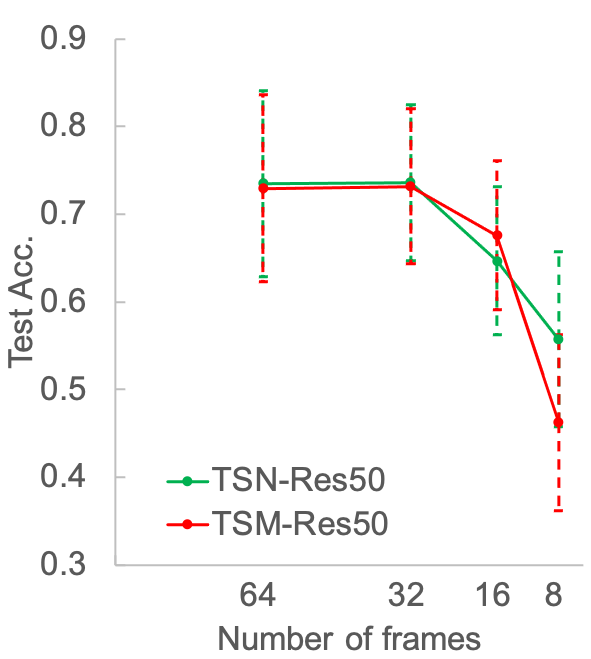}
%
\vspace{-1em}
\caption{The performance of TSM-Res50 and TSN-Res50 with different values of $K$ on PPMI(left)  and ADNI(right) datasets.}
\label{fig2}
\vspace{-1.5em}
\end{figure}

\vspace{-2mm}
\wqz{There are hundreds of magnetic resonance images for a patient and it is difficult to feed all images into a deep model due to the limitation of computation resources. A common way is to draw $K$ images as input, thus the value of $K$ is crucial for the final performance. In this section, we conduct ablation studies on $K$. Fig. \ref{fig2} presents the performance of TSM-Res50 and TSN-Res50 on PPMI and ADNI with different values of $K$. Obviously, a large $K$ slows down the training and inference processes and a small $K$ could reduce the recognition accuracy. We can see that for both TSM-Res50 and TSN-Res50, $K=32$ is a better choice, which achieves much higher accuracy than using $K=16$ and $K=8$. Though using $K=64$ can achieve similar accuracy on ADNI, the computational complexity doubles. In this paper, we suggest using $K=32$ to balance effectiveness and efficiency.}

\vspace{-3mm}
\subsection{Ablation study on data augmentation}\vspace{-2mm}

\wqz{Data augmentation plays a vital role in existing video recognition frameworks. Particularly, if the training set is small, data augmentation is useful for mitigating overfitting. In this section, we investigate the effects of using different data augmentation strategies --- random rotation, CutMix and MixUp which are popular techniques in video recognition. Table~\ref{tab3} shows the performance of TSM-Res50 with different data augmentation strategies on PPMI and ADNI. We can easily make the conclusion that data augmentation can significantly boost the performance, e.g., the performance on PPMI surges from 0.793 to 0.866. Interestingly, applying MixUp hurts the performance, e.g., TSM+rotate achieves 0.832 and 0.659 on PPMI and ADNI, while the performance of TSM+rotate+MixUp decreases to 0.813 and 0.516 respectively. The possible reason is that magnetic resonance images are similar and the directly weighted sum of two images could make the lesion areas less distinctive, leading to misclassification. In contrast, CutMix remarkably improves the performance, since it crops an image area and the lesion areas are still distinguishable.}


\vspace{-2mm}
\section{Conclusion}\vspace{-2mm}
\label{sec:majhead}

\wqz{In this paper, we have investigated MRI classification using advanced video recognition frameworks, finding that (1)~directly applying advanced video recognition frameworks, such as TSM and SlowFast performance significantly better than the widely used 3D models, (2)~TSM is a better choice for MRI classification when considering both effectiveness and efficiency, (3)~video recognition techniques like data augmentation are useful for MRI classification. Possible future directions include leveraging advanced \emph{pre-training}~\cite{liao2022muscle} \& \emph{fine-tuning strategies}~\cite{li2019delta} to solve MRI tasks and explaining the diagnosis resultss~\cite{li2022interpretable,li2022interpretdl} delivered by MRI models.}


\bibliographystyle{IEEEbib}
\bibliography{refs}

\begin{thebibliography}{10}

\bibitem{8211486}
K~A N N~P Gunawardena, R~N Rajapakse, and N~D Kodikara,
\newblock ``Applying convolutional neural networks for pre-detection of
  alzheimer's disease from structural mri data,''
\newblock in {\em M2VIP}, 2017, pp. 1--7.

\bibitem{lecun2015deep}
Yann LeCun, Yoshua Bengio, and Geoffrey Hinton,
\newblock ``Deep learning,''
\newblock {\em Nature}, vol. 521, no. 7553, pp. 436--444, 2015.

\bibitem{li2020rifle}
Xingjian Li, Haoyi Xiong, Haozhe An, Cheng-Zhong Xu, and Dejing Dou,
\newblock ``Rifle: Backpropagation in depth for deep transfer learning through
  re-initializing the fully-connected layer,''
\newblock in {\em ICML}, 2020, pp. 6010--6019.

\bibitem{lin2019tsm}
Ji~Lin, Chuang Gan, and Song Han,
\newblock ``Tsm: Temporal shift module for efficient video understanding,''
\newblock in {\em ICCV}, 2019, pp. 7083--7093.

\bibitem{he2016deep}
Kaiming He, Xiangyu Zhang, Shaoqing Ren, and Jian Sun,
\newblock ``Deep residual learning for image recognition,''
\newblock in {\em CVPR}, 2016, pp. 770--778.

\bibitem{marek2011parkinson}
Kenneth Marek, Danna Jennings, Shirley Lasch, Andrew Siderowf, Caroline Tanner,
  Tanya Simuni, Chris Coffey, Karl Kieburtz, Emily Flagg, Sohini Chowdhury,
  et~al.,
\newblock ``The parkinson progression marker initiative (ppmi),''
\newblock {\em Progress in neurobiology}, vol. 95, no. 4, pp. 629--635, 2011.

\bibitem{9288068}
Tahjid~Ashfaque Mostafa and Irene Cheng,
\newblock ``Parkinson’s disease detection using ensemble architecture from mr
  images,''
\newblock in {\em BIBE}, 2020, pp. 987--992.

\bibitem{zhang2022mining}
Jing Zhang,
\newblock ``Mining imaging and clinical data with machine learning approaches
  for the diagnosis and early detection of parkinson’s disease,''
\newblock {\em npj Parkinson's Disease}, vol. 8, no. 1, pp. 1--15, 2022.

\bibitem{kushol2022addformer}
Rafsanjany Kushol, Abbas Masoumzadeh, Dong Huo, Sanjay Kalra, and Yee-Hong
  Yang,
\newblock ``Addformer: Alzheimer’s disease detection from structural mri
  using fusion transformer,''
\newblock in {\em ISBI}. IEEE, 2022, pp. 1--5.

\bibitem{li2022trans}
Chao Li, Yue Cui, Na~Luo, Yong Liu, Pierrick Bourgeat, Jurgen Fripp, and Tianzi
  Jiang,
\newblock ``Trans-resnet: Integrating transformers and cnns for alzheimer’s
  disease classification,''
\newblock in {\em ISBI}. IEEE, 2022, pp. 1--5.

\bibitem{dhinagar20213d}
Nikhil~J Dhinagar, Sophia~I Thomopoulos, Conor Owens-Walton, Dimitris
  Stripelis, Jose~Luis Ambite, Greg Ver~Steeg, Daniel Weintraub, Philip Cook,
  Corey McMillan, and Paul~M Thompson,
\newblock ``3d convolutional neural networks for classification of
  alzheimer’s and parkinson’s disease with t1-weighted brain mri,''
\newblock in {\em SIPAIM}. SPIE, 2021, vol. 12088, pp. 277--286.

\bibitem{liang2021alzheimer}
Gongbo Liang, Xin Xing, Liangliang Liu, Yu~Zhang, Qi~Ying, Ai-Ling Lin, and
  Nathan Jacobs,
\newblock ``Alzheimer’s disease classification using 2d convolutional neural
  networks,''
\newblock in {\em EMBC}. IEEE, 2021, pp. 3008--3012.

\bibitem{oh2020identifying}
Jihoon Oh, Baek-Lok Oh, Kyong-Uk Lee, Jeong-Ho Chae, and Kyongsik Yun,
\newblock ``Identifying schizophrenia using structural mri with a deep learning
  algorithm,''
\newblock {\em Frontiers in psychiatry}, vol. 11, pp. 16, 2020.

\bibitem{xiong2022grod}
Haoyi Xiong, Ruosi Wan, Jian Zhao, Zeyu Chen, Xingjian Li, Zhanxing Zhu, and
  Jun Huan,
\newblock ``Grod: Deep learning with gradients orthogonal decomposition for
  knowledge transfer, distillation, and adversarial training,''
\newblock {\em ACM Transactions on Knowledge Discovery from Data (TKDD)}, vol.
  16, no. 6, pp. 1--25, 2022.

\bibitem{wu2022survey}
Fei Wu, Qingzhong Wang, Jiang Bian, Ning Ding, Feixiang Lu, Jun Cheng, Dejing
  Dou, and Haoyi Xiong,
\newblock ``A survey on video action recognition in sports: Datasets, methods
  and applications,''
\newblock {\em IEEE Transactions on Multimedia}, 2022.

\bibitem{carreira2017quo}
Joao Carreira and Andrew Zisserman,
\newblock ``Quo vadis, action recognition? a new model and the kinetics
  dataset,''
\newblock in {\em CVPR}, 2017, pp. 6299--6308.

\bibitem{Hara_2017_ICCV}
Kensho Hara, Hirokatsu Kataoka, and et~al.,
\newblock ``Learning spatio-temporal features with 3d residual networks for
  action recognition,''
\newblock in {\em ICCV Workshops}, 2017.

\bibitem{simonyan2014two}
Karen Simonyan and Andrew Zisserman,
\newblock ``Two-stream convolutional networks for action recognition in
  videos,''
\newblock {\em NIPS}, vol. 27, 2014.

\bibitem{wang2016temporal}
Limin Wang, Yuanjun Xiong, Zhe Wang, Yu~Qiao, Dahua Lin, Xiaoou Tang, and Luc
  Van~Gool,
\newblock ``Temporal segment networks: Towards good practices for deep action
  recognition,''
\newblock in {\em ECCV}. Springer, 2016, pp. 20--36.

\bibitem{feichtenhofer2019slowfast}
Christoph Feichtenhofer, Haoqi Fan, Jitendra Malik, and Kaiming He,
\newblock ``Slowfast networks for video recognition,''
\newblock in {\em ICCV}, 2019, pp. 6202--6211.

\bibitem{zhao2021efficient}
Baoxin Zhao, Haoyi Xiong, Jiang Bian, Zhishan Guo, Cheng-Zhong Xu, and Dejing
  Dou,
\newblock ``Como: Efficient deep neural networks expansion with convolutional
  maxout,''
\newblock {\em IEEE Transactions on Multimedia}, vol. 23, pp. 1722--1730, 2021.

\bibitem{yun2020videomix}
Sangdoo Yun, Seong~Joon Oh, Byeongho Heo, Dongyoon Han, and Jinhyung Kim,
\newblock ``Videomix: Rethinking data augmentation for video classification,''
\newblock {\em arXiv preprint arXiv:2012.03457}, 2020.

\bibitem{yun2019cutmix}
Sangdoo Yun, Dongyoon Han, Seong~Joon Oh, Sanghyuk Chun, Junsuk Choe, and
  Youngjoon Yoo,
\newblock ``Cutmix: Regularization strategy to train strong classifiers with
  localizable features,''
\newblock in {\em ICCV}, 2019, pp. 6023--6032.

\bibitem{zhang2017mixup}
Hongyi Zhang, Moustapha Cisse, Yann~N Dauphin, and David Lopez-Paz,
\newblock ``mixup: Beyond empirical risk minimization,''
\newblock {\em arXiv preprint arXiv:1710.09412}, 2017.

\bibitem{he2015delving}
Kaiming He, Xiangyu Zhang, Shaoqing Ren, and Jian Sun,
\newblock ``Delving deep into rectifiers: Surpassing human-level performance on
  imagenet classification,''
\newblock in {\em ICCV}, 2015, pp. 1026--1034.

\bibitem{srivastava2014dropout}
Nitish Srivastava, Geoffrey Hinton, Alex Krizhevsky, Ilya Sutskever, and Ruslan
  Salakhutdinov,
\newblock ``Dropout: a simple way to prevent neural networks from
  overfitting,''
\newblock {\em JMLR}, vol. 15, no. 1, pp. 1929--1958, 2014.

\bibitem{liao2022muscle}
Weibin Liao, Haoyi Xiong, Qingzhong Wang, and et~al.,
\newblock ``Muscle: Multi-task self-supervised continual learning to pre-train
  deep models for x-ray images of multiple body parts,''
\newblock in {\em MICCAI}. Springer, 2022, pp. 151--161.

\bibitem{li2019delta}
Xingjian Li, Haoyi Xiong, Hanchao Wang, Yuxuan Rao, Liping Liu, and Jun Huan,
\newblock ``Delta: Deep learning transfer using feature map with attention for
  convolutional networks,''
\newblock in {\em ICLR}, 2019.

\bibitem{li2022interpretable}
Xuhong Li, Haoyi Xiong, Xingjian Li, Xuanyu Wu, Xiao Zhang, Ji~Liu, Jiang Bian,
  and Dejing Dou,
\newblock ``Interpretable deep learning: Interpretation, interpretability,
  trustworthiness, and beyond,''
\newblock {\em Knowledge and Information Systems}, vol. 64, no. 12, pp.
  3197--3234, 2022.

\bibitem{li2022interpretdl}
Xuhong Li, Haoyi Xiong, Xingjian Li, Xuanyu Wu, Zeyu Chen, and Dejing Dou,
\newblock ``Interpretdl: Explaining deep models in paddlepaddle,''
\newblock {\em Journal of Machine Learning Research}, vol. 23, no. 197, pp.
  1--6, 2022.

\end{thebibliography}

\end{document}